\documentclass[conference, 10pt]{IEEEtran}
\IEEEoverridecommandlockouts

\usepackage{cite}
\usepackage{amsmath,amssymb,amsfonts}
\usepackage{algorithmic}
\usepackage{graphicx}
\usepackage{textcomp}
\usepackage{xcolor}
\usepackage{subcaption}
\usepackage[letterpaper, top=0.75in, bottom=1.1in, left=0.70in, right=0.70in]{geometry}

\def\BibTeX{{\rm B\kern-.05em{\sc i\kern-.025em b}\kern-.08em
    T\kern-.1667em\lower.7ex\hbox{E}\kern-.125emX}}

\begin{document}
\setlength{\columnsep}{0.25in}

\makeatletter
\def\@maketitle{%
\begin{center}
    \large \textbf{Note: This paper has been accepted for oral presentation at the WF-IoT 2025 conference. The version posted here will be removed once the final paper is published on the IEEE website.} 
\end{center}
\vspace{0.2in} 
\begin{center}%
{\LARGE \@title \par}%
\vskip 1em
{\large
\begin{tabular}[t]{c}
\@author
\end{tabular}\par}%
\vskip 1em
\end{center}%
}
\makeatother

\title{Garbage Vulnerable Point Monitoring\\ using IoT and Computer Vision}

\author{
    \IEEEauthorblockN{R. Kumar\textsuperscript{}, A. Lall\textsuperscript{}, S. Chaudhari\textsuperscript{}, M. Kale,\textsuperscript{} A. Vattem \textsuperscript{}}
    \IEEEauthorblockA{\textit{\textsuperscript{1}International Institute of Information Technology-Hyderabad (IIIT-H), India}}
    \IEEEauthorblockA{\textit{\textsuperscript{2}United Way of Hyderabad, Hyderabad, India}}
    \IEEEauthorblockA{
    \{rahul.ku, anuradha.vattem\}@research.iiit.ac.in, ayush.lall@alumni.iiit.ac.in \\ madhuvanti@unitedwayhyderabad.org, sachin.c@iiit.ac.in}
}

\maketitle
\begin{abstract}
This paper proposes a smart way to manage municipal solid waste by using the Internet of Things (IoT) and computer vision (CV) to monitor illegal waste dumping at garbage vulnerable points (GVPs) in urban areas. The system can quickly detect and monitor dumped waste using a street-level camera and object detection algorithm. Data was collected from the Sangareddy district in Telangana, India. A series of comprehensive experiments was carried out using the proposed dataset to assess the accuracy and overall performance of various object detection models. Specifically, we performed an in-depth evaluation of YOLOv8, YOLOv10, YOLO11m, and RT-DETR on our dataset. Among these models, YOLO11m achieved the highest accuracy of 92.39\% in waste detection, demonstrating its effectiveness in detecting waste. Additionally, it attains a mAP@50 of 0.91, highlighting its high precision. These findings confirm that the object detection model is well-suited for monitoring and tracking waste dumping events at GVP locations. Furthermore, the system effectively captures waste disposal patterns, including hourly, daily, and weekly dumping trends, ensuring comprehensive daily and nightly monitoring. 


\end{abstract}
\section{Introduction}
Waste is a growing environmental issue around the world \cite{bilalMicrobialBioremediationRobust2020}. Waste management is a global challenge, and many governments struggle with it due to limited infrastructure and capacity for collection, processing, and treatment. Despite its complexity, involving both technical and human factors, the problem's urgency has drawn significant attention from researchers, industries, and governments to develop sustainable solutions across the entire waste management system.

Garbage vulnerable points (GVPs) are locations where waste is dumped without proper disposal. These sites produce foul odors, serve as breeding grounds for pests and stray animals, cause soil and water pollution, and create unsanitary conditions in localities \cite{shahEvaluatingGarbageVulnerable2023}. Cleaning GVPs requires additional resources, increasing the costs of waste management. Some residents resort to burning trash, which worsens air pollution and harms public health. Stray animals further scatter waste, worsening the situation and diminishing neighborhood cleanliness. 

As cities grow, managing waste efficiently becomes a significant challenge. To build cleaner and smarter cities, we need modern technologies that support sustainable development \cite{leaInternetThingsArchitects2018}. By integrating the Internet of Things (IoT) and Computer Vision (CV) technologies, waste collection and disposal can become more efficient, affordable, and environmentally sustainable. IoT devices can give real-time updates, so waste is collected only when needed, saving time, fuel, and money. IoT and CV help create an innovative and efficient waste management system that reduces costs and keeps cities cleaner.

There has been some work on technology solutions for various aspects of waste management. The work in \cite{shahEvaluatingGarbageVulnerable2023} studied that several smart cities in India use IoT technologies to improve waste management. It describes a low-cost, energy-efficient sensor node using a single-chip microcontroller and ultrasonic sensor to monitor real-time bin fill levels.
In \cite{cerchecciLowPowerIoT2018}, design a low-cost, energy-efficient IoT sensor node for smart city waste management. It checks bin fill levels and a LoRa module for long-range, low-power data transfer. Additionally, the authors in 
\cite{dealmeidaLoRaWANInfrastructureUrban2023} presents the feasibility of using a LoRaWAN communication infrastructure for implementing the differentiated waste collection application at the University of Campinas (Unicamp) in Brazil. Through network simulations, the research evaluates its effectiveness while integrating other smart campus applications to assess the network’s overall capabilities and limitations. The work in \cite{shengInternetThingsBased2020} highlights the integration of LoRa communication with a TensorFlow-based deep learning model for smart waste management. The system uses object detection to classify waste into separate compartments controlled by servo motors. It includes GPS tracking and personnel identification via RFID. Work in \cite{haqueIoTBasedEfficient2020} describes the IoT-based waste collection system with smart bins that enable real-time monitoring. It features an enhanced navigation system to identify the most efficient routes for collection. In a simulated scenario, it reduces travel distance by 30.76\% compared to traditional methods. The study in \cite{voskergianSmartEwasteManagement2023} shows three object detection models—YOLOv5s, YOLOv7-tiny, and YOLOv8s—for e-waste detection. Among them, YOLOv8s performed the best, with an mAP@50 of 72\% and an mAP@50-95 of 52\%. In \cite{shahabSolidWasteManagement2022}, the authors address key challenges in solid waste management, including the unavailability of web platforms for public engagement, insufficient real-time tracking of waste bins and collection vehicles, and widespread instances of unauthorized waste disposal. Few studies have considered using technology to monitor illegal waste dumping and mitigate GVPs. The study in \cite{rahmanIntelligentWasteManagement2022} evaluates waste management solutions using deep learning and IoT. A CNN model is used to classify waste as digestible or indigestible with 91.31\% accuracy. The system includes a smart bin with sensors and a microcontroller, using IoT for real-time monitoring and Bluetooth for short-range access through a mobile app.

Note that the existing literature\cite{shahEvaluatingGarbageVulnerable2023, cerchecciLowPowerIoT2018,dealmeidaLoRaWANInfrastructureUrban2023,shengInternetThingsBased2020,haqueIoTBasedEfficient2020,voskergianSmartEwasteManagement2023,shahabSolidWasteManagement2022} focuses on general waste management, and none of the work is on GVP monitoring, which is the focus of this paper. The key contributions of this paper are as follows:
\begin{itemize}
    \item A cost-effective system integrating IoT and CV is proposed to detect illegal waste dumping in real-time. 
    \item The method was evaluated using a dataset of over 19,280 images captured by a camera at a GVP location in Sangareddy, Telangana, India, over two months. \footnote{The collected data will be made publicly available for future research.}
    \item Comparative evaluation of different object detection models such as YOLOv8, YOLOv10, YOLO11m, and RT-DETR is carried out for the proposed solution. 
     \item Different analyses are carried out on the collected data to understand hourly, daily, and weekday patterns related to dumping and cleaning the GVP.
\end{itemize}

The paper is organized as follows: Section II provides information on the methodology, which is further divided into two subsections: IoT-based camera setup and dataset collection campaign. Section III discusses the CV-based models, which are further distributed into three subsections: end-to-end pipeline for training and detection, evaluation metrics, and computational setup. Section IV presents the results and discussion, covering five subsections: waste disposal behavior by hour, daily waste patterns, weekday analysis of waste dumping activities, and cost analysis. Lastly, Section V concludes the paper.
\section{METHODOLOGY}
A camera deployed on the streets of Sangareddy, Telangana, India, collects data on waste disposal behaviors. The approach entails data collection and preparation to train an object detection model for waste detection. Once operational, the model analyzes test samples to identify waste, aiding in tracking disposal trends and optimizing GVP site management.


\subsection{IoT-Based camera setup}
Designed for versatile deployment, the EZVIZ H8c 4G camera \cite{ezvizinc.EZVIZH8c4G} supports 4G LTE connectivity using a nano SIM card and includes an RJ45 Ethernet interface to ensure a stable and secure wired connection. Featuring 2K resolution and Color Night Vision, it delivers high-quality video in any lighting. Its weatherproof design ensures durability in diverse conditions, while Active Defense with a siren and strobe light helps deter intruders. The camera offers a panoramic view and supports local storage of up to 512 GB microSD and cloud storage via EZVIZ CloudPlay, making it a robust residential and commercial security option.

\subsection{Data collection campaign}
At the Sangareddy location, a street-side camera was installed to capture live video and monitor waste dumping activities. The dataset was collected under real-world conditions to assess the practical performance of both the camera and the model for effective deployment. Over two months, the camera operated continuously, recording 24 hours daily at a frame rate of 30 fps. The recorded video was converted into frames to optimize data processing, with one frame extracted every five minutes. The video was converted into frames, with one captured every five minutes. This five-minute interval was chosen to reduce the cumulative waste data and minimize computational costs. This process resulted in more than 17,280 frames from the camera. Five thousand frames were annotated using LabelImg and Roboflow image annotator tools. During annotation, bounding boxes were drawn around regions of interest and labeled as either waste or non-waste to train the model for accurate detection. These annotations were saved in a format compatible with YOLO, ensuring a smoother and faster workflow and eliminating the need to convert the data during runtime. The annotated data was divided into two sets: 80\% for training the model and 20\% for testing the model. Data augmentation techniques were applied to the training set to improve the model's performance. In the dataset, 2,000 images were flipped to create more variations because it helps the model generalize better and become more robust. By seeing objects and patterns from different angles, the model learns to recognize them more accurately, no matter their orientation. This is especially useful when the position of objects keeps changing, like in tasks such as object detection and 
 behavior analysis. After augmentation, the total number of images increased to 19,280. This process helps the model learn better by exposing it to more diverse examples, improving its ability to detect objects accurately in different scenarios. 
 \newcommand{\figuresize}{0.49\textwidth}  



\section{CV based  models}
Among existing object detection models, YOLO is widely acknowledged for its fast inference and high accuracy. To evaluate the performance of various models, a thorough comparison was conducted using our dataset, including YOLOv8, YOLOv10, YOLO11m, and RT-DETR. 
\begin{figure*}[h]
  \captionsetup[subfigure]{belowskip=+5pt}
  \centering
  \begin{subfigure}{\textwidth}
    \centering
    \includegraphics[width=0.8\textwidth]{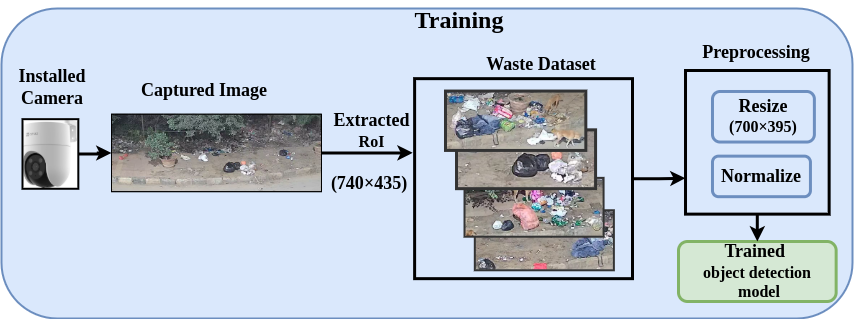}
    \caption{The YOLO11m training pipeline captures images from an installed camera, extracts regions of interest (RoI) showing waste, and creates a specialized dataset. The images are resized to $700 \times 395$  and normalized for consistency. This processed dataset trains the object detection model, ensuring accurate and robust waste detection.}
    \label{subfig: Training}
  \end{subfigure}
  \hfill
  \begin{subfigure}{\textwidth}
    \centering
    \includegraphics[width=0.8\textwidth]{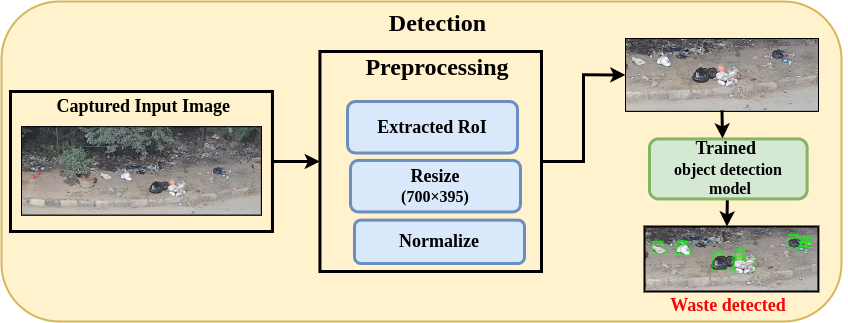}
    \caption{The detection pipeline extracts the region of interest (RoI) from captured images, preprocesses them, and detects contours around waste objects. The YOLO11m model then segments and detects the waste and counts the number of waste present at GVP locations.}
    \label{subfig: detection}
  \end{subfigure}
  \caption{Automating the end-to-end process for training and object detection}
  \label{fig:Architecture}
\end{figure*}

YOLOv8 \cite{vargheseYOLOv8NovelObject2024}  is a fast and versatile model excelling in object detection, segmentation, and classification. It improves upon YOLOv5 with a CSPNet backbone for enhanced feature extraction, an FPN+PAN neck for multi-scale detection, and an anchor-free approach. Additionally, its unified Python package and CLI streamline model training and deployment.

YOLOv10 \cite{wangYOLOv10RealTimeEndtoEnd2024a} is an efficient real-time object detector that improves the model architecture and the post-processing steps in the detection pipeline. It introduces a unique approach called consistent dual assignments for NMS-free training, which simplifies and speeds up the detection process by eliminating the need for traditional non-maximum suppression (NMS). It uses a well-balanced model design strategy focused on accuracy and efficiency, leading to better performance with lower latency. 

YOLO11 \cite{khanamYOLOv11OverviewKey2024} is a high-performance object detection model balancing speed and accuracy for real-time applications. It enhances feature extraction through the C2PSA mechanism and C3K2 block, improving spatial analysis and small object detection without compromising inference speed. The SPFF module optimizes efficiency, making YOLOv11 well-suited for real-time object recognition.

RT-DETR \cite{zhaoDETRsBeatYOLOs2024} is a transformer-based model and the first real-time, end-to-end object detector. It uses the power of transformers for computer vision tasks, making object detection faster and more efficient. It gives flexible speed tuning by adjusting the number of decoder layers, allowing it to adapt to different scenarios without retraining. However, this flexibility has a higher computational power demand than YOLO models. YOLO11, on the other hand, is built for speed and efficiency, making it faster and less computationally intensive.

\subsection{End-to-End Pipeline for Training and Detection}

 Fig. \ref{subfig: Training} shows the training pipeline used to build the waste detection model that enables real-time detection of waste dumping. The process begins by collecting images of waste using installed cameras. These images are then converted into frames, and an ROI is applied to extract relevant portions, creating a waste dataset. All images are resized to a uniform number of $700 \times 395$ pixels to maintain consistency, and pixel values are normalized to speed up and stabilize training. The processed images are fed into the YOLO11 fast and accurate object detection model. During the training time of the model, to enhance the robustness of the model, both daytime and nighttime images were included during training. The model's parameters were optimized using Stochastic Gradient Descent (SGD) with a momentum of 0.937 and a learning rate of 0.01, aiming to minimize the discrepancy between predicted and ground truth bounding boxes and class labels. Data loading was parallelized using eight worker threads (\texttt{num\_workers=8}) to ensure efficient throughput and optimal GPU utilization. The training process was carried out over 100 epochs, with data augmentation techniques—such as blurring the images and horizontal flipping—employed to improve the model’s generalization capabilities. Following training, the model was evaluated on a separate test set to assess its performance. 
 
Fig. \ref{subfig: detection} shows that the detection process begins with waste images captured at the GVP's location. First, image preprocessing (like extracting the ROI, resizing, and normalizing) ensures the input matches the format the object detection model was trained on: $700 \times 395$ pixels. After completing the preprocessing step, the trained model scans the images in real-time, using its learned parameters to identify waste objects at the GVP locations. It predicts bounding boxes around detected waste and assigns labels based on patterns it recognizes during training. Contour detection is used to analyze object shapes and edges to enhance accuracy, enabling better boundary definition. The model identifies ROI-specific areas in the image most likely to contain waste, optimizing computational resource utilization. This system marks detected waste with bounding boxes, delivering actionable insights for waste management. This efficient pipeline ensures fast and accurate detection, even in cluttered environments, while incorporating illegal waste dumping analysis techniques to adapt to real-world conditions.
\subsection{Evaluation metrics}
This section covers key evaluation metrics, including precision, recall, F1-score, mAP@50, and accuracy. The mean Average Precision (mAP) evaluates how accurately an object detection model predicts and classifies bounding boxes. It uses Intersection over Union (IoU) to compare predicted and ground truth boxes, with predictions considered correct if IoU exceeds a set threshold, typically 0.50 (mAP@50). mAP captures model performance by balancing precision and recall, considering true positives, false positives, and false negatives. Precision measures the proportion of correctly predicted objects among all predicted objects, indicating the reliability of predictions. Recall assesses the model’s ability to detect all relevant objects, calculated as the ratio of true positives to actual positives. The F1-score, the harmonic mean of precision and recall, provides a balanced evaluation, especially useful when a class imbalance exists or when false positives and false negatives carry different costs. Accuracy measures the overall correctness of predictions by dividing the number of correct predictions by the total number of predictions made.
\subsection{Computational setup}
The experiment was conducted on a system equipped with an NVIDIA RTX 2080 Ti GPU with 12 GB of dedicated memory and 128 GB of RAM. The testing environment utilizes Ubuntu 22.04 LTS as the operating system. The proposed YOLO11m models were trained using PyTorch Build Stable (2.5.0) with CUDA 12.6 support.
\section{Results and Discussion}

The collected waste disposal data was used to compare the object detection models presented in the previous subsection. Behavioral patterns at various temporal resolutions—hourly, daily, and weekly scales—are also analyzed, followed by hardware cost analysis. All the results have been discussed in the upcoming subsections

\subsection{Comparison of object detection models}

\begin{table}[t]
    \caption{Comparison of different object detection models on the GVP dataset.}
    \centering
    \renewcommand{\arraystretch}{1.2} 
    \begin{tabular}{|c|c|c|c|c|}
        \hline
        \textbf{Metric}   & \textbf{YOLOv8m} & \textbf{YOLOv10m} & \textbf{RT-DETR} & \textbf{YOLO11m}  \\
        \hline
        Model Size & 102 MB & 88 MB & 66 MB & \textbf{74 MB}  \\
        \hline
        Precision  & 0.91   & 0.89   & 0.82   & \textbf{0.94}  \\
        \hline
        Recall & 0.84   & 0.81   & 0.79   & \textbf{0.84} \\
        \hline
        F1-Score   & 0.87   & 0.84   & 0.80   & \textbf{0.88}  \\
        \hline
        mAP@50     & 0.87   & 0.86   & 0.84   & \textbf{0.91} \\
        \hline
        Accuracy   & 82.63\% & 86.34\% & 84.24\% & \textbf{92.39\%} \\
        \hline
    \end{tabular}
    \label{tab:model-comparison}
\end{table}
Table \ref{tab:model-comparison} compares different object detection models based on their precision, recall, F1-score, and mAP50 on the waste detection dataset. Among these, YOLO11m outperforms all other models, achieving the highest mAP@50 of 0.91 and an accuracy of 92.39\%, highlighting its superior capability in accurately detecting waste within the RoI. It also records the highest precision of 0.94, indicating a strong ability to minimize false positives and correctly identify waste. In comparison, YOLOv10m also performs well, with a mAP@50 of 0.86 and an accuracy of 86.34\%, making it the second most effective model. RT-DETR, with a smaller model size of 66 MB, achieves moderate performance (mAP@50 of 0.84 and accuracy of 84.24\%) and can be considered a lightweight alternative when computational efficiency is prioritized. YOLOv8m, although slightly larger in size (102 MB), delivers balanced results with a mAP@50 of 0.87 and an accuracy of 82.63\%. This reliability positions the object detection model as an effective tool for tracking waste disposal activities and analyzing behavioral patterns at the identified location. Given that YOLO11m exhibited the highest accuracy among the considered methods, it is considered for the rest of the results section.

\subsection{Waste disposal behavior by hour}

\begin{figure}[h]
  \centering  \includegraphics[width=0.88\linewidth]{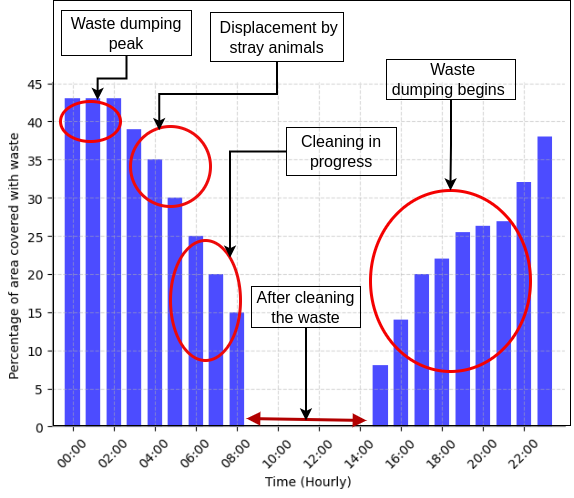}
  \caption{The hourly waste coverage within the Region of Interest (RoI) at the GVP was computed as an average over the entire data collection timeframe.}
  \label{fig: waste available}
\end{figure}

\begin{figure}[h]
  \centering  
  \includegraphics[width=0.88\linewidth]{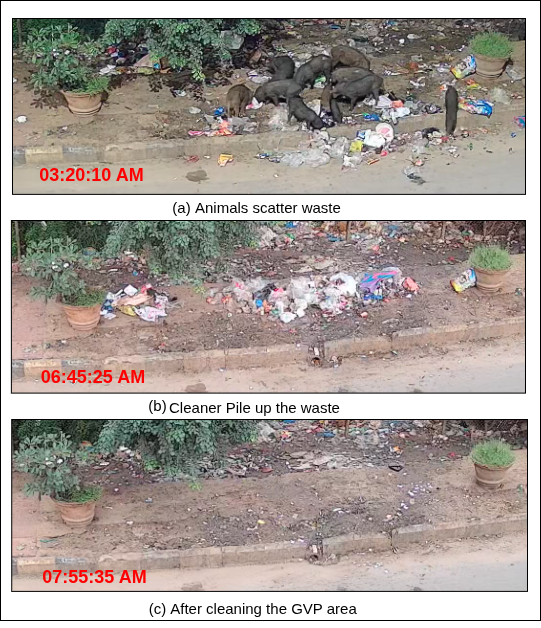}
  \caption{Different activities affecting the waste distribution at the GVP.}
  \label{fig: Area_picture}
\end{figure}
Fig. \ref{fig: waste available} shows the hourly trend of waste accumulation at the GVP, highlighting key patterns throughout the day. It can be seen that the GVP is covered with the largest amount of waste between midnight and 3 AM, with 39\% ROI covered. After that, the values decrease as stray animals scatter the waste (or consume food waste) out of the RoI between 3 AM and 6 AM, as can be seen from Fig. \ref{fig: waste available}(a). Sanitation workers typically arrive between 6:00 and 8:00 AM and clean the area in two phases. In the first phase, one set of workers collects and piles up the waste into a single location, as shown in Fig. \ref{fig: Area_picture}(b). This leads to overlapping waste items, reducing the waste area. In the second phase, a garbage truck completely picks up this piled waste, making the area clean, as shown in Fig. \ref{fig: Area_picture} (c). The location is clean for the next 7-8 hours. The dumping again starts slowly at 3 PM, increasing between 8 PM and 11 PM, followed by a peak between 11 PM and midnight. This pattern shows that nearby commercial establishments may dump waste after closing their shops.

\subsection{Daily waste patterns}

\begin{figure}[tbh]
  \centering
  \includegraphics[width=0.88\linewidth]{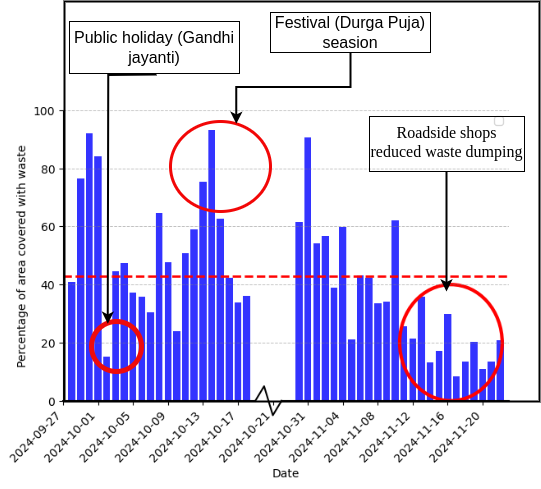}
  \caption{Daily average percentage of ROI covered with waste at the GVP for the two months.}
  \label{fig:day-wise}
\end{figure}

\subsection{Weekday analysis of waste dumping activities}

\begin{figure}[tbh]
  \centering
  \includegraphics[width=0.88\linewidth]{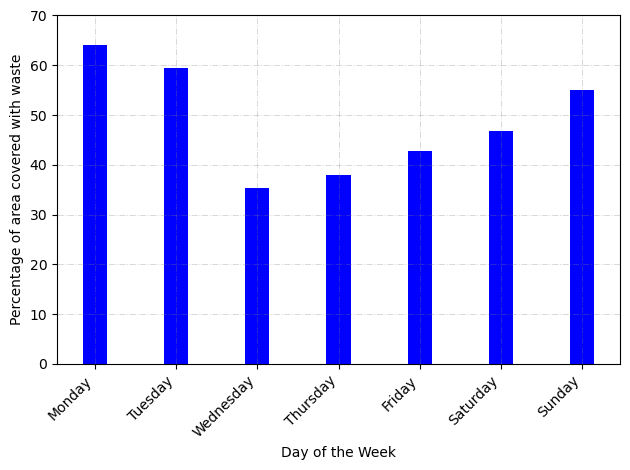}
  \caption{Weekly distribution of waste at GVP}
  \label{fig: weekday}
\end{figure}

Fig. \ref{fig: weekday} shows the average weekday waste area over two months. The analysis highlights distinct waste disposal patterns at GVP locations, with noticeable variations across different days of the week. Waste accumulation is significantly higher on Mondays and Tuesdays, likely due to the buildup over the weekend. A recurring pattern of public holidays on Wednesdays results in a noticeable drop in waste disposal. Waste levels rise again on Thursdays and Fridays before experiencing a slight decline on Saturdays. Understanding these trends enables authorities to optimize waste management strategies and allocate more resources.
\subsection{Hardware cost analysis}

\begin{table}[h]
    \caption{Evaluating the economic feasibility of the low-cost setup}
    \centering
     \renewcommand{\arraystretch}{1.2}
    \begin{tabular}{|c|c|c|c|}
        \hline
        \textbf{Components} & \textbf{Quantity} & \textbf{Cost (INR)} \\
        \hline
        Camera & 1 &  4266\\
        \hline
        SanDisk Ultra 256GB memory card & 1  &  1699 \\
        \hline
        Cost of Labour  &  1  & 500\\
        \hline
        Miscellaneous cost  & -  & 500 \\
        \hline
        \textbf{Total}  & - &  \textbf{7,315} \\
        \hline
    \end{tabular}
    \label{tab:cost-analysis}
\end{table}
Table \ref{tab:cost-analysis} presents a cost analysis of the proposed low-cost setup designed for data collection, waste detection, and behavior analysis at the GVP location. The estimated one-time cost is INR 7,015, with a monthly recurring recharge cost of INR 300. This budget-friendly approach aims to enhance waste management efficiency while keeping operational costs low, making it a scalable solution for multiple locations.
\section{Conclusion}
This paper explored how CV and IoT technologies can enhance real-time waste detection and monitoring, particularly at GVPs in urban fringe areas. Utilizing camera-based data collection, the system captured waste disposal behaviors, monitored waste accumulation in real-time, and identified peak dumping times on hourly, daily, and weekly scales. A comparative analysis was conducted using several object detection models, including YOLOv8, YOLOv10, YOLO11m, and RT-DETR, with YOLO11m demonstrating the highest performance in terms of accuracy. Consequently, YOLO11m was deployed for real-time detection and monitoring of illegal dumping activities. The system achieved a waste detection accuracy of 92.39\% and a mean average precision at 0.5 (mAP@50) score of 0.91 across the entire dataset. By leveraging temporal analysis of disposal patterns, the proposed system supports municipalities in optimizing waste collection schedules and strengthening the enforcement of anti-dumping regulations. This data-driven approach addresses key challenges in urban waste management, improving operational efficiency, reducing costs, and providing valuable insights for city planners and regulatory bodies, ultimately contributing to cleaner urban environments and enhanced community well-being. Future work will focus on deploying cameras across multiple GVP locations to enable more comprehensive data collection and behavioral analysis.
\section*{Acknowledgements}
This research was supported by the Smart City Research Center Lab and partially funded by the Qualcomm Edge AI Lab at IIIT Hyderabad, with no conflict of interest. The authors thank the Sangareddy Municipality for their support in data collection.
\bibliographystyle{ieeetr}
\bibliography{references}
\vspace{12pt}

\end{document}